\pdfoutput=1

\documentclass[11pt]{article}

\usepackage[]{ACL2023}

\usepackage{times}
\usepackage{latexsym}
\usepackage{graphicx} 
\usepackage{amsmath}
\usepackage{amssymb}
\usepackage{colortbl}
\usepackage{xcolor}
 
\usepackage[T1]{fontenc}

\usepackage[utf8]{inputenc}
\usepackage{tipa}

\usepackage{microtype}

\usepackage{inconsolata}

\usepackage{comment}
\usepackage{paralist}
\usepackage{multirow}

\usepackage[table]{xcolor}

\definecolor{improve}{RGB}{198, 239, 206} 
\definecolor{worse}{RGB}{255, 199, 206}   

\newcommand\blfootnote[1]{%
  \begingroup
  \renewcommand\thefootnote{}\footnote{#1}%
  \addtocounter{footnote}{-1}%
  \endgroup
}

\title{myMediWhisper: Construction of Burmese Medical Speech Corpus and Whisper Fine-Tuning for Clinical Dialogue ASR}

\author{
  \textbf{Ye Kyaw Thu}\textsuperscript{1,2,$\dagger$,*},
  \textbf{Ye Bhone Lin}\textsuperscript{2,3,$\dagger$,$\ddagger$},
  \textbf{Thura Aung}\textsuperscript{2,4,$\dagger$,*}, \\
  \textbf{Htet Arkar}\textsuperscript{4,$\ddagger$},
  \textbf{Myat Oo Swe}\textsuperscript{4,$\ddagger$}, 
  \textbf{Thet Htet San}\textsuperscript{4,$\ddagger$}, \\
  \textbf{Min Thiha Tun}\textsuperscript{2,$\ddagger$},
  \textbf{Thazin Myint Oo}\textsuperscript{2}, 
  \textbf{Thepchai Supnithi}\textsuperscript{1,*} \\
  \textsuperscript{1}National Electronics and Computer Technology Center (NECTEC), Thailand \\
  \textsuperscript{2}Language Understanding Laboratory, Myanmar \\
  \textsuperscript{3}King Mongkut's University of Technology Thonburi, Thailand \\
  \textsuperscript{4}King Mongkut's Institute of Technology Ladkrabang, Thailand
}

\begin{document}
\maketitle

\blfootnote{\textsuperscript{*}Corresponding authors.}
\blfootnote{\textsuperscript{$\dagger$}These authors contributed equally to this work.}
\blfootnote{\textsuperscript{$\ddagger$}Work done during internship at LU. Lab., Myanmar.}

\begin{abstract}
Although Whisper models benefit from large-scale multilingual pre-training, their performance on Burmese medical speech remains limited. This work presents a Burmese medical speech recognition framework built on a high-quality 28-hour corpus recorded and validated by native speakers. We fine-tune Whisper models using full fine-tuning (FFT) and parameter-efficient fine-tuning (PEFT) with LoRA. To evaluate robustness, we apply waveform- and spectrogram-level data augmentation under controlled noise and simulated room acoustics. While augmentation reduces performance on clean speech, it significantly improves robustness in noisy and reverberant environments across FFT and PEFT settings. Our best-performing system, fully fine-tuned myMediWhisper-Medium without augmentation, achieves a state-of-the-art Word Error Rate (WER) of 23.44\%, outperforming much larger general-domain fine-tuned models. Dataset and other resources can be found at the Huggingface repository: \url{https://huggingface.co/datasets/LULab/mediTalk-mm-rdy}.
\end{abstract}

\begin{figure*}[t]
    \centering
    \includegraphics[width=0.9\linewidth]{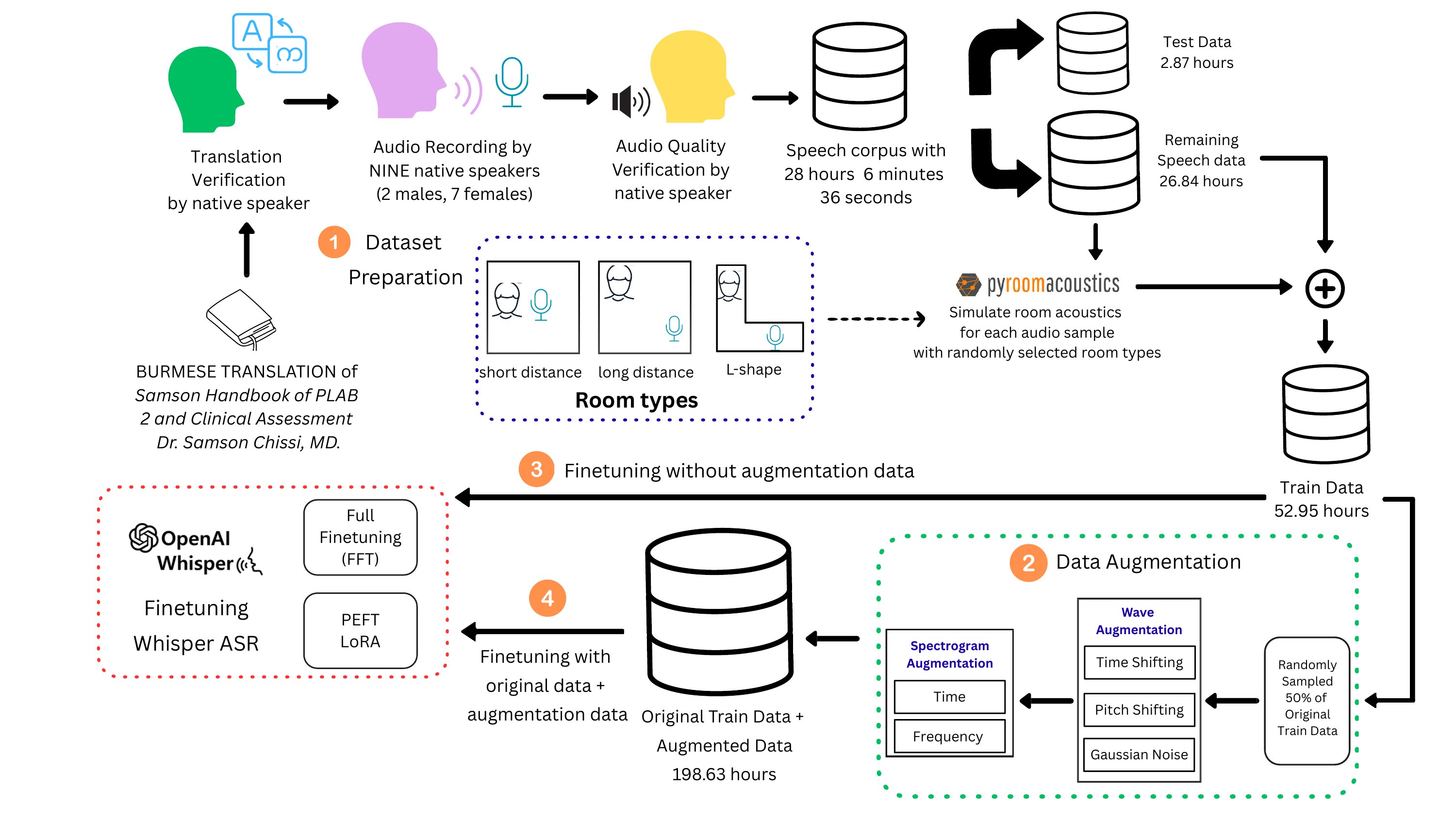}
    \caption{Pipeline for (1) dataset preparation, (2) augmentation, fine-tuning (3) without and (4) with augmentation}
    \vspace{-5mm}
    \label{fig:pipeline}
\end{figure*}

\section{Introduction}

Automatic Speech Recognition (ASR) has become increasingly important in the medical domain, enabling the conversion of spoken interactions into text, which supports clinical documentation and reduces the workload of healthcare professionals \cite{ASR_health_record_2010}. ASR systems can also be used for tasks such as voice-based data entry, information retrieval, and clinical workflow support. However, previous studies show that ASR performance in medical settings remains challenging, with reported accuracies ranging from 78\% to 92\% \cite{GOSS2019103938, Ajami_2016}.

Compared to high-resource languages such as English, research on Burmese ASR remains limited \cite{HayMarSoeNaing2015, AyeNyeinMon2019, KhaingZarMon2019}, mainly due to the lack of publicly available speech corpora \cite{oo-etal-2020-burmese}. In the medical domain, available resources are even more scarce. The myMediCon corpus \cite{mymedicon} is one of the few existing Burmese medical speech datasets and reports the best baseline word error rate (WER) of 19.0\% using an RNN-based ASR model. However, the speech data have not yet been publicly released. 

\noindent \textbf{Proposed Experiments.}
To address the issues mentioned above, we construct a Burmese medical speech corpus using native-speaker recordings and incorporate simulated room acoustics during training data preparation to reflect realistic clinical conditions. We release the dataset publicly and establish finetuned Whisper \cite{radford2022robustspeechrecognitionlargescale} ASR models. Models are fine-tuned with and without waveform- and spectrogram-level data augmentation. The evaluation is conducted on clean test data, with robustness further analyzed under controlled noise levels and simulated room environments.

\noindent \textbf{Proposed Studies.}
Using this setting, we examine previously underexplored aspects of Burmese medical ASR, focusing on model adaptation and robustness in low-resource conditions.
\begin{compactenum}[\bf RQ1]
    \item \textbf{Fine-tuning Strategies.}
    \emph{How do full fine-tuning (FFT) and parameter-efficient fine-tuning (PEFT) of Whisper models compare for Burmese medical ASR?}
    
    \item \textbf{Effect of Data Augmentation.}
    \emph{How does data augmentation influence recognition accuracy and robustness?}
    
    \item \textbf{Robustness to Acoustic Variability.}
    \emph{How well do our models generalize to noise and room acoustic variations when evaluated?}
\end{compactenum}

\noindent \textbf{Contributions.}
We contribute (i) a publicly available Burmese medical speech corpus, (ii) Whisper-based ASR benchmarks using FFT and PEFT, and (iii) a systematic analysis of robustness under simulated noise, and room acoustic variations. 

\section{Related Work}
\noindent \textbf{Burmese ASR.}
Previous Burmese ASR research has focused on a limited set of domains. Early work trained GMM- and DNN-based acoustic models on manually constructed travel and conversational speech corpora \cite{HayMarSoeNaing2015}. Later studies introduced general-domain datasets covering daily conversations and news, reporting domain-dependent ASR performance \cite{Mon2017WebNewsASR, Mon2018MyanmarCNN, AyeNyeinMon2019}. More recently, myMediCon \cite{mymedicon} presented one of the first medical-domain Burmese speech corpora and evaluated end-to-end ASR models based on RNN and Transformer architectures. However, the dataset remains limited in scale, and prior work does not investigate large-scale pretrained models or robustness under acoustic variations. The speech datasets have also not been publicly released, limiting reproducibility and further benchmarking. In contrast, our work focuses on fine-tuning Whisper models on a newly constructed medical speech corpus and systematically analyzing robustness using data augmentation and room acoustic simulation. 

\noindent \textbf{Whisper fine-tuning.}
Previous studies have shown that Whisper can be fine-tuned for low-resource language adaptation and gained lower WER than vanilla pretrained multilingual Whisper models \cite{_zyilmaz_2025, aung-etal-2024-thonburian, pranida-etal-2025-asr}. Despite improvements, for morphologically complex languages like Swahili, challenges remain \cite{sharma-etal-2025-fine}. In case of low-resource adaptation, \cite{gete2025whisperingamharicfinetuningwhisper} shows that mixing specialized Amharic datasets with the FLEURS dataset \cite{fleurs2022arxiv} reinforces performance better than training on new data alone. 

\section{Background}
\label{sec:background}
\subsection{Whisper Architecture}
\label{subsec:whisper_arch}
Whisper \cite{radford2022robustspeechrecognitionlargescale} is a sequence-to-sequence encoder-decoder Transformer model engineered for unified speech processing tasks, including multilingual automatic speech recognition (ASR), translation, and voice activity detection. The acoustic pipeline ingests an 80-dimensional log-Mel spectrogram extracted from 30-second audio segments, denoted as $\mathbf{X} = [\mathbf{x}_1, \mathbf{x}_2, \dots, \mathbf{x}_T] \in \mathbb{R}^{80 \times T}$, where $T$ represents the temporal frame dimension. 

The audio encoder processes $\mathbf{X}$ through convolutional layers and consecutive Transformer blocks to yield a sequence of continuous hidden representations $\mathbf{H}$. The text decoder subsequently generates token predictions $\hat{y}_t$ auto-regressively, conditioned upon the preceding tokens $\hat{y}_{1:t-1}$, special task prompts $\mathbf{p}$ (e.g., language and task identifiers), and the encoder representations $\mathbf{H}$ via cross-attention. Formally, this sequence-to-sequence transformation is defined as:
\begin{equation}
    \mathbf{H} = \text{AudioEncoder}(\mathbf{X})
\end{equation}
\begin{equation}
    \hat{y}_t = \text{TextDecoder}(\mathbf{p}, \hat{y}_{1:t-1}, \mathbf{H})
\end{equation}

\subsection{Rank-Stabilized Low-Rank Adaptation (rsLoRA)}
\label{subsec:rslora_method}
To fine-tune the massive parameter space of Whisper models under strict hardware limitations, we employ Parameter-Efficient Fine-Tuning (PEFT) via Low-Rank Adaptation (LoRA) \cite{hu2022lora}, specifically leveraging Rank-Stabilized LoRA (rsLoRA) \cite{kalajdzievski2023rankstabilizationscalingfactor}. Consider a targeted projection layer with a pre-trained, frozen weight matrix $\mathbf{W}_0 \in \mathbb{R}^{d_1 \times d_2}$. Traditional LoRA decomposes the weight update $\Delta \mathbf{W}$ into two trainable low-rank matrices $\mathbf{B} \in \mathbb{R}^{d_1 \times r}$ and $\mathbf{A} \in \mathbb{R}^{r \times d_2}$, parameterized by an intrinsic rank $r \ll \min(d_1, d_2)$.

While standard LoRA scales the adapter output by a static factor $\frac{\alpha}{r}$, rsLoRA introduces a modified scaling factor $\frac{\alpha}{\sqrt{r}}$ to stabilize the learning dynamics when scaling up to high-rank configurations. The forward propagation for a given input vector $\mathbf{x}$ is formulated as:
\begin{equation}
    \mathbf{h} = \mathbf{W}_0 \mathbf{x} + \Delta \mathbf{W} \mathbf{x} = \mathbf{W}_0 \mathbf{x} + \frac{\alpha}{\sqrt{r}} \mathbf{B}\mathbf{A}\mathbf{x}
\end{equation}

In our architecture setup, we deliberately apply this adaptation to the critical multi-head attention components, isolating the query and value projection matrices ($\mathbf{W}_q, \mathbf{W}_v \in \{\text{\texttt{q\_proj}}, \text{\texttt{v\_proj}}\}$). We employ an expanded low-rank capacity ($r = 128$) scaled via $\alpha = 256$ combined with a dropout regularizer of $0.05$ applied directly to $\Delta \mathbf{W}$ to capture highly specialized medical dictation while completely freezing $\mathbf{W}_0$.

\section{Experiment Setup}
\label{sec:exp_setup}

As illustrated in Figure~\ref{fig:pipeline}, the speech dataset preparation consists of four main stages:
(1) transcript data preparation through translation verification,
(2) audio recording and audio quality verification, (3) training data construction with acoustic simulation, and (4) data augmentation. The dataset is then used to finetune Whisper ASR models with both FFT and PEFT strategies.

\noindent \textbf{Transcript Data Preparation.}
For transcript preparation, we adopt the Burmese translations of the \textit{Samson Handbook of PLAB 2 and Clinical Assessment} by Dr. Samson Chissi, MD, originally released for machine translation research in \cite{ei-san-etal-2022-improving}.
The source material comprises structured medical scenarios and clinical dialogues designed for PLAB 2 OSCE preparation \cite{samson2015}.
All translated Burmese sentences were manually verified by two native Burmese speakers fluent in English to ensure semantic accuracy, medical consistency, and linguistic naturalness.
Following verification, a total of 14,517 sentences were retained and used as transcription data for subsequent speech recording.

\noindent \textbf{Audio Recording and Quality Verification.}
The speech corpus was collected from nine native Burmese speakers (two male and seven female), all university students in their early twenties who can read Burmese transliterated medical terminologies correctly. To reflect realistic data collection conditions while maintaining accessibility and scalability, recordings were conducted using built-in microphones on personal devices (smartphones and laptops) in quiet environments at a sampling rate of 16~kHz. Each speaker was provided with pre-verified medical transcripts to ensure consistency across recordings.
Given that Burmese is a syllabic and tonal language in which syllable-level variations can change meaning \cite{multicsd_burmese, Mon2017TonesCNN}, all recorded utterances were manually reviewed and recordings that did not strictly match the transcripts at the syllable level were discarded.
After quality verification, the resulting corpus contains 28 hours, 6 minutes, and 36 seconds of high-quality Burmese medical speech suitable for clinical dialogue ASR.

\noindent \textbf{Training Data Construction.}
From the verified corpus, 52.95 hours of speech were allocated as training data, with 2.87 hours reserved for evaluation.
To improve robustness to real-world acoustic conditions, room acoustics were simulated for each training audio sample using the \texttt{Pyroomacoustics}\footnote{\url{https://pyroomacoustics.readthedocs.io/}} Python library \cite{scheibler2018pyroomacoustics}.
Room impulse responses were generated by randomly sampling different room types, including L-shaped, random-distance configurations. The simulated impulse responses were combined with clean speech signals to create reverberant training samples, enabling the model to better generalize to diverse recording environments. Table~\ref{table:speaker} summarizes the per-speaker statistics of the corpus.

\noindent \textbf{Data Augmentation.}
To increase diversity and improve model robustness, a structured multi-stage data augmentation strategy was applied to the training data.
First, 50\% of the original training set was randomly sampled for augmentation.
On this sampled subset, three waveform-level augmentation \cite{audiomentations2023} methods were applied independently: random time shifting, pitch shifting, and additive Gaussian noise.
This stage generates augmented data equivalent to three times the sampled subset. Each waveform-augmented sample was then further processed using two spectrogram-level augmentation methods \cite{park2019specaugment}, namely time masking with a probability of 0.3, masking 10 consecutive time steps, and frequency masking with a probability of 0.1 across 64 frequency bands.
Both spectrogram augmentations were applied to each waveform-augmented variant, further increasing data diversity.
This augmentation process follows a factor of $0.5 \times 3 \times 2$ and expands the train data from 52.95 hours to 198.63 hours, corresponding to an approximate 3.75$\times$ increase.

\noindent \textbf{Hyperparameter Setup} For both the training and evaluation phases, we configured the model with a batch size of 4. The optimization process was executed over a single epoch with a learning rate of \(1\times10^{-5}\). To circumvent the local storage limitations of the cloud environment and optimize data throughput, we implemented a streaming data pipeline. All experiments were conducted within a cloud-hosted Kaggle Notebooks environment, leveraging distributed parallel computing across dual NVIDIA Tesla T4 GPUs (providing a combined \(32\text{ GB}\) of VRAM) to efficiently handle synchronous operations.

\begin{table*}[ht!]
\renewcommand{\arraystretch}{1}
\centering
\footnotesize
\begin{tabular}{ccrrrrrr}
\hline
\textbf{Speaker ID} & \textbf{Gender} &
\multicolumn{2}{c}{\textbf{Train}} &
\multicolumn{2}{c}{\textbf{Test (Evaluation)}} &
\multicolumn{2}{c}{\textbf{Total}} \\
\cline{3-4} \cline{5-6} \cline{7-8}
& & \textbf{\# Utt.} & \textbf{Dur. (hrs)} &
  \textbf{\# Utt.} & \textbf{Dur. (hrs)} &
  \textbf{\# Utt.} & \textbf{Dur. (hrs)} \\
\hline
sp01 & Male & 7,364 & 14.62 & 410 & 0.78 & 7,774 & 15.40 \\
sp02 & Female & 1,800 & 5.39  & 100 & 0.30 & 1,900 & 5.69  \\
sp03 & Female & 1,800 & 3.63  & 100 & 0.20 & 1,900 & 3.83  \\
sp04 & Female & 5,400 & 8.96  & 300 & 0.49 & 5,700 & 9.45  \\
sp05 & Female & 900  & 1.83  & 50  & 0.10 & 950  & 1.93  \\
sp06 & Female & 3,600 & 7.06  & 200 & 0.42 & 3,800 & 7.48  \\
sp07 & Female & 2,700 & 5.70  & 150 & 0.32 & 2,850 & 6.02  \\
sp08 & Female & 1,800 & 3.31  & 100 & 0.20 & 1,900 & 3.51  \\
sp09 & Male & 900  & 1.45  & 50  & 0.08 & 950  & 1.53  \\
\hline
\multicolumn{2}{c}{\textbf{Total}} & 20,264 & 52.95 & 1,460 & 2.87 & 21,724 & 55.82 \\
\hline
\end{tabular}
\caption{Per-speaker statistics of the corpus, including train, test, and total utterances and durations after adding simulated speech with different room acoustics (Figure \ref{fig:pipeline}).}
\label{table:speaker}
\vspace{-3mm}
\end{table*}

\begin{table*}[ht!]
\renewcommand{\arraystretch}{1}
\footnotesize
\centering
\definecolor{topfirst}{RGB}{198, 239, 206}  
\definecolor{topsecond}{RGB}{217, 234, 211} 
\definecolor{topthird}{RGB}{242, 242, 242}  

\newcommand{\best}[1]{\cellcolor{topfirst}\textbf{#1}}
\newcommand{\second}[1]{\cellcolor{topsecond}#1}
\newcommand{\third}[1]{\cellcolor{topthird}#1}

\resizebox{\textwidth}{!}{%
\begin{tabular}{lrrrrrrr}
\hline
\textbf{Model} & \textbf{Param} & \textbf{SER} $\downarrow$ & \textbf{DER} $\downarrow$ & \textbf{IER} $\downarrow$ & \textbf{WER} $\downarrow$ & \textbf{chrF} $\uparrow$ & \textbf{RTF} $\downarrow$ \\
\hline

\multicolumn{8}{l}{\textbf{Zero-shot ASR Baselines}} \\
sil-ai/wav2vec2-bloom-speech-mya & 300M & 92.16 & \third{3.41} & 4.35 & 98.53 & 0.1523 & \best{0.015} \\
chuuhtetnaing/whisper-large-v3-myanmar & 1550M & 23.61 & \second{2.26} & 4.82 & 32.03 & 0.7140 & 1.187 \\
facebook/MMS-1B & 1000M & 31.87 & \best{1.98} & 3.51 & 37.90 & 0.6362 & \second{0.044} \\

\hline

\multicolumn{8}{l}{\textbf{Zero-shot Whisper Models}} \\

Vanilla Whisper Tiny & 39M & 54.59 & 45.41 & 135.20 & 235.2 & 1e-16 & \third{0.097} \\
Vanilla Whisper Base & 74M & 84.68 & 15.32 & 504.19 & 604.19 & 1e-16 & 0.116 \\
Vanilla Whisper Small & 244M & 65.13 & 34.87 & 193.79 & 293.79 & 1e-16 & 0.721 \\
Vanilla Whisper Medium & 769M & 75.68 & 24.32 & 156.07 & 256.07 & 1e-16 & 0.323 \\
Vanilla Whisper Large v2 & 1550M & 48.92 & 51.08 & 49.16 & 149.16 & 1.61e-4 & 0.946 \\

\hline
\multicolumn{8}{l}{\textbf{PEFT myMediWhisper Models (Ours)*}} \\

myMediWhisper Tiny & 1.6M & 66.00 & 22.95 & 42.96 & 115.68 & 0.1678 & 0.172 \\
myMediWhisper Base & 3.1M & 76.03 & 10.35 & 17.49 & 99.57 & 0.2169 & 0.340 \\
myMediWhisper Small & 8.4M & 32.18 & 29.41 & 1.08 & 64.80 & 0.4274 & 0.357 \\
myMediWhisper Medium & 25.2M & 35.10 & 8.30 & 5.23 & 43.52 & 0.5839 & 0.408 \\
myMediWhisper Large v2 & 40.6M & 27.22 & 11.32 & 1.40 & 41.57 & 0.6283 & 1.006 \\

myMediWhisper Tiny Aug & 1.6M & 45.90 & 44.76 & 11.79 & 92.28 & 0.1331 & 0.132 \\
myMediWhisper Base Aug & 3.1M & 81.49 & 12.05 & 13.42 & 118.21 & 0.1841 & 0.370 \\
myMediWhisper Small Aug & 8.4M & 51.26 & 30.49 & 2.31 & 84.71 & 0.2436 & 0.371 \\
myMediWhisper Medium Aug & 25.2M & 49.83 & 15.40 & 5.61 & 72.72 & 0.3918 & 0.821 \\
myMediWhisper Large v2 Aug & 40.6M & 33.80 & 9.58 & 2.62 & 47.63 & 0.5736 & 1.056 \\
\hline
\multicolumn{8}{l}{\textbf{FFT myMediWhisper Models (Ours)}} \\

myMediWhisper Tiny & 39M & 35.57 & 13.93 & 2.73 & 52.53 & 0.5283 & 0.148 \\
myMediWhisper Base & 74M & 18.84 & 12.17 & \second{0.82} & 31.83 & 0.7144 & 0.204 \\
myMediWhisper Small & 244M & \third{13.56} & 11.37 & \third{0.88} & \third{25.81} & \third{0.7779} & 0.347 \\
myMediWhisper Medium & 769M & \best{11.52} & 11.15 & \best{0.77} & \best{23.44} & \second{0.7988} & 0.691 \\

myMediWhisper Tiny Aug & 39M & 61.57 & 17.92 & 4.14 & 83.63 & 0.2667 & 0.144 \\
myMediWhisper Base Aug & 74M & 34.33 & 13.20 & 3.78 & 51.31 & 0.5562 & 0.197 \\
myMediWhisper Small Aug & 244M & 20.20 & 12.25 & 2.08 & 34.53 & 0.7154 & 0.361 \\
myMediWhisper Medium Aug & 769M & \second{12.05} & 10.91 & 1.77 & \second{24.73} & \best{0.8052} & 0.696 \\
\hline
\end{tabular}
}
\caption{Comparison of ASR performance for zero-shot Whisper models, fully fine-tuned myMediWhisper models, and parameter-efficient fine-tuned (PEFT) myMediWhisper models on the myMediWhisper evaluation dataset. Metric rankings are highlighted column-wise as \protect\colorbox{topfirst}{\textbf{Best}}, \protect\colorbox{topsecond}{2nd Best}, and \protect\colorbox{topthird}{3rd Best}. *For PEFT models, the reported parameter count refers only to the trainable adapter parameters.}
\label{tab:asr_results}
\vspace{-5mm}
\label{tab:asr_results}
\end{table*}

\section{Results and Discussion}
\label{sec:results}

\subsection{ASR Performance Comparison}
\label{subsec:asr_performance}

\noindent \textbf{Comparison against External ASR Baselines.}
To contextualize the performance of our proposed models, we compare them against established external baselines in Table~\ref{tab:asr_results}. Massively Multilingual Speech (\texttt{facebook/MMS-1B}) \cite{pratap2023mms} struggles significantly in this specialized domain, yielding a high Word Error Rate (WER) of 37.90\%. Similarly, general-domain fine-tuned models exhibit performance gaps when encountering our evaluation set. While the open-source fine-tuned \texttt{whisper-large-v3-myanmar} \cite{ chuuhtetnaing_2024_whisper_large_v3_myanmar} achieves a baseline WER of 32.03\% and \texttt{sil-ai/wav2vec2-bloom-speech-mya} \cite{sil_global_ai_2022_wav2vec2_bloom_speech_mya} , our FFT domain-specific models comfortably outperform it at a fraction of the computational scale. For instance, \texttt{myMediWhisper-Small} (244M) achieves a lower WER of 25.81\% while reducing the parameter footprint by over 84\%, illustrating that general-domain fine-tuning on open assets is insufficient for localized Burmese medical terminology and clinical dialogue.

\noindent \textbf{Effectiveness of FFT and PEFT.}
We compare FFT and PEFT to evaluate their effectiveness under low-resource and hardware-constrained conditions. As shown in Table~\ref{tab:asr_results}, domain adaptation leads to massive accuracy gains over vanilla zero-shot Whisper variants. When computationally feasible, FFT consistently achieves the lowest error rates, yielding relative WER reductions ($\Delta$\%) of up to 94.73\% for Whisper-Base (dropping from 604.19\% to 31.83\%) and 90.85\% for Whisper-Medium (dropping from 256.07\% to 23.44\%) without acoustic augmentation. These improvements are primarily driven by large reductions in substitution errors (SER) and deletion errors (DER), while insertion errors (IER) remain well-controlled.

\noindent \textbf{Scalability under Memory Constraints.}
Despite its strong performance, FFT is not practical for larger models in our setup, as vanilla Whisper-Large-v2 results in out-of-memory (OOM) errors during full backpropagation on our hardware stack. In contrast, configuring PEFT via Rank-Stabilized LoRA (\texttt{use\_rslora=True}) enables stable, resource-conscious adaptation of Large-v2. Our configuration targets the primary attention projections (\texttt{["q\_proj", "v\_proj"]}) using an expanded rank configuration ($r=128$, $\alpha=256$) and a dropout rate of $0.05$. This targeted adaptation unlocks the training of Large-v2, achieving a WER reduction of 72.13\% without augmentation (WER = 41.57\%) and 68.07\% with data augmentation (WER = 47.63\%) as shown in Table~\ref{tab:asr_results}. While PEFT generally underperforms FFT when comparing identical smaller model sizes—particularly yielding higher substitution rates—it serves as a critical scaling mechanism to activate large parameter architectures under strict hardware restrictions.

\noindent \textbf{Accuracy and Efficiency.}
As shown across the configurations in Table~\ref{tab:asr_results}, PEFT models exhibit higher real-time factors (RTF) than their corresponding FFT counterparts, indicating an inference latency penalty. For example, PEFT increases the RTF from 0.204 to 0.340 for Whisper-Base, and from 0.691 to 0.408 for unaugmented Whisper-Medium, with the augmented variant jumping further to 0.821. This decoding overhead is directly tied to our low-rank parameter choices; maintaining a high rank ($r=128$) across the multi-head attention modules preserves high representation capacity for complex medical dialogues, but introduces a non-trivial matrix multiplication overhead during frozen sequential generation. These results highlight the multi-dimensional trade-offs between word accuracy, memory footprint during training, and operational throughput. Overall, in response to \textbf{RQ1}, FFT provides the strongest localized adaptation when hardware resources permit, while our PEFT configuration successfully unlocks large-model scaling for finetuning under limited hardware at the expense of higher inference latency.



\begin{figure}[t!]
    \centering
    \includegraphics[width=\linewidth]{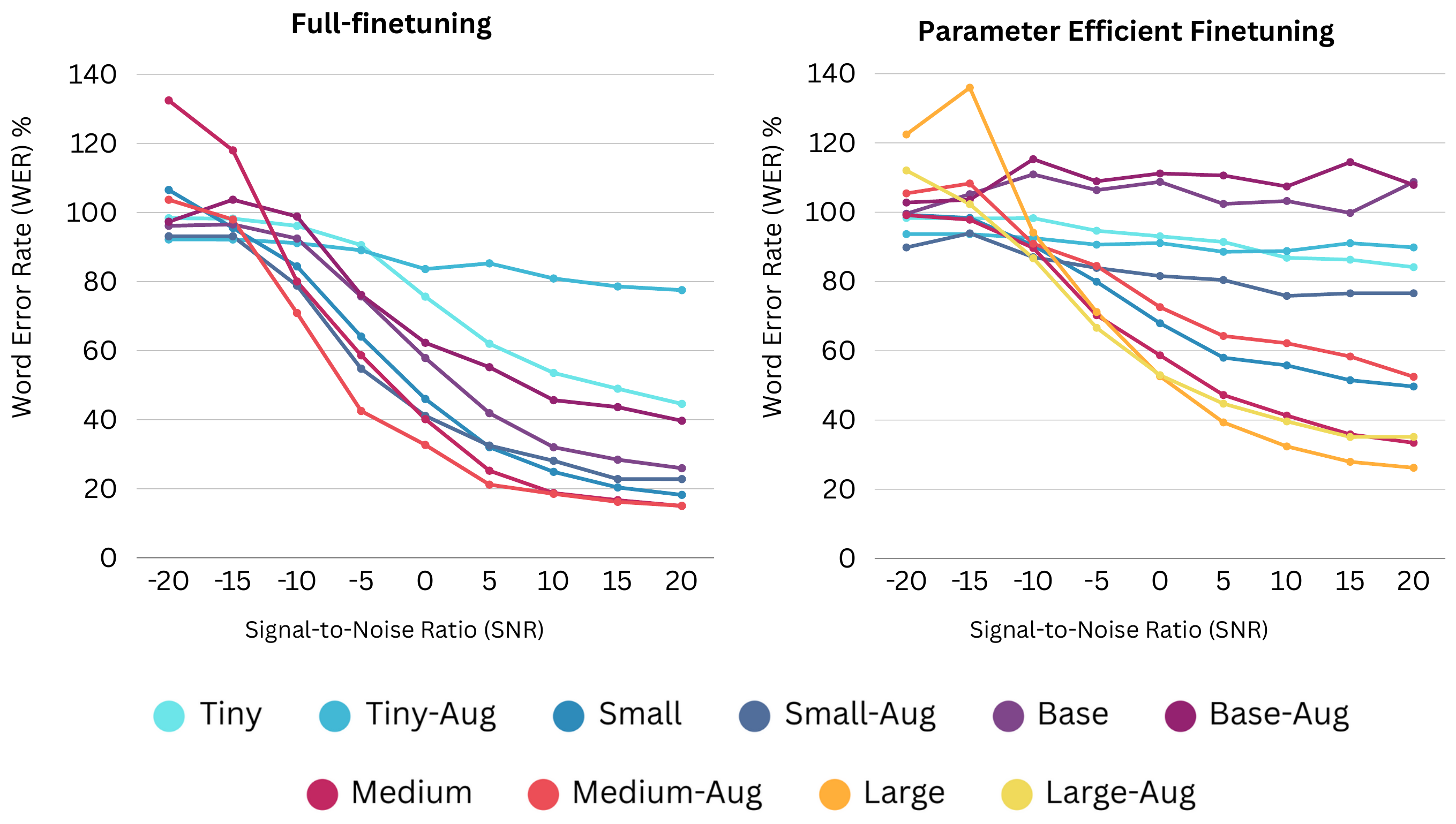}
    \caption{WER under varying Signal-to-Noise Ratio (SNR) conditions for myMediWhisper models trained with and without data augmentation.}
    \label{fig:robustness_snr}
\end{figure}

\begin{figure}[t!]
    \centering
    \includegraphics[width=\linewidth]{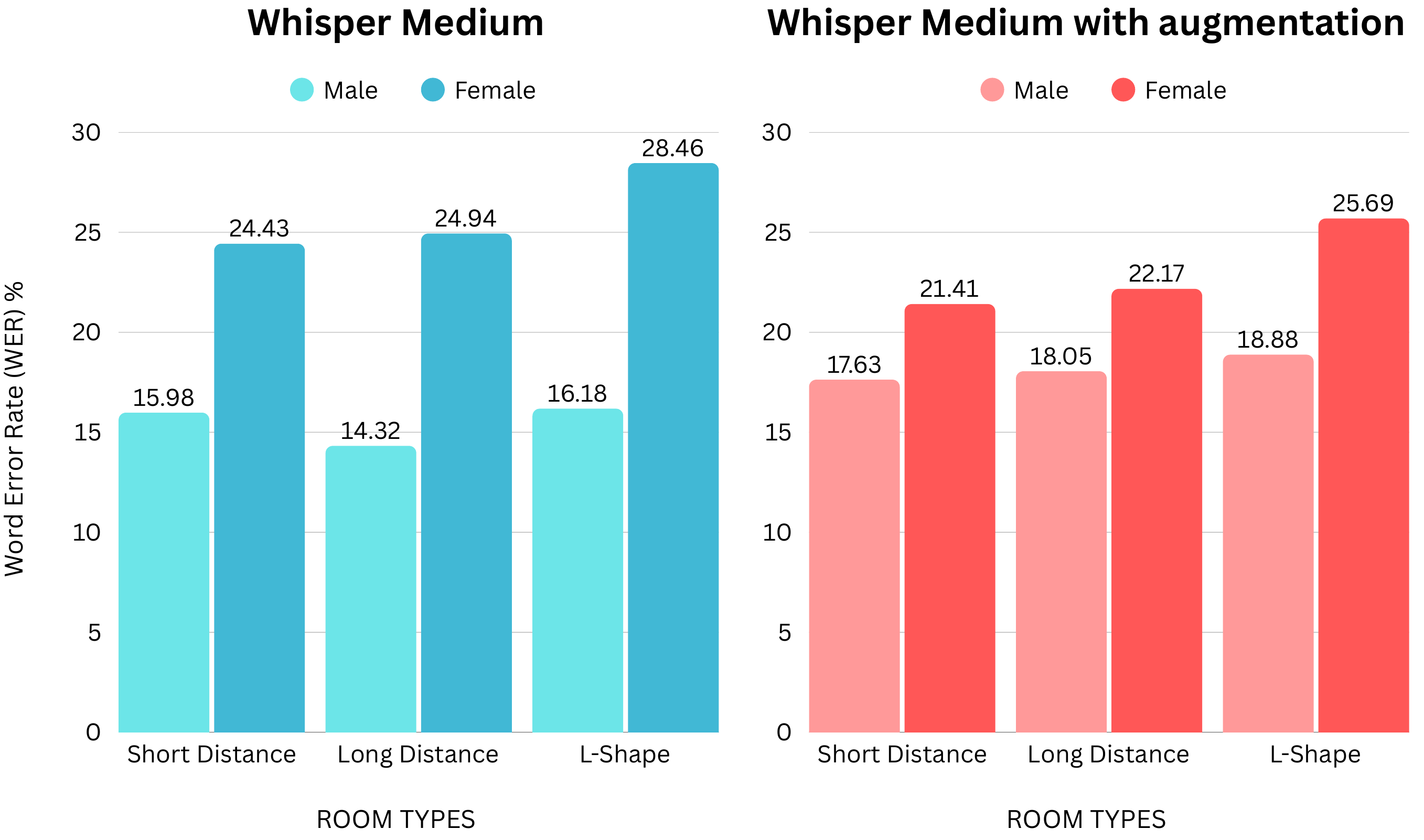}
    \caption{Effect of simulated room acoustics on WER for Whisper-Medium, comparing without and with augmentation across room types and speaker gender.}
    \label{fig:wer_room_types}
    \vspace{-5mm}
\end{figure}

\subsection{Robustness to Acoustic Variation}
\label{subsec:acoustic_robustness}

In real clinical environments, ASR systems are exposed to background noise and reverberation that are not present in clean training data. Evaluating robustness to such acoustic variability is therefore essential for medical ASR, particularly in low-resource settings. To this end, we analyze the effects of data augmentation and acoustic conditions on model performance, focusing on room acoustic generalization and noise robustness.

\noindent \textbf{Robustness to SNR Conditions.} We first evaluate robustness to background noise by testing all Whisper models under varying SNR conditions, reporting WER for models trained with and without data augmentation (\textbf{RQ2}). Figure~\ref{fig:robustness_snr} shows that data augmentation generally improves robustness at lower SNR levels for medium and larger models, while smaller models benefit less consistently and may degrade under severe noise. This suggests that sufficient model capacity is required to effectively leverage augmented acoustic variability.

\noindent \textbf{Robustness to Room Acoustics.} We then examine robustness to room acoustics using Whisper-Medium trained with both FFT and PEFT. We sample 40 utterances (20 male, 20 female) and simulate three room types with different spatial characteristics (Figure~\ref{fig:pipeline}). As shown in Figure~\ref{fig:wer_room_types}, WER increases under long-distance and L-shaped room configurations, reflecting more challenging reverberation conditions. Performance differences between male and female speakers remain relatively small across room types, indicating limited gender bias in the evaluated setting.

Overall, these results indicate that fine-tuned Whisper models generalize reasonably well to unseen room acoustics and noisy conditions, and that data augmentation primarily benefits noise robustness in larger models, addressing robustness and bias considerations (\textbf{RQ3}).

\section{Error Analysis}
\label{sec:error_analysis}

To evaluate the fine-grained phonetic and structural failure modes of myMediWhisper Medium Aug, which is the most robust model (Section \ref{subsec:acoustic_robustness}), we conducted an error analysis using sequence alignment at the syllable level. 

\begin{figure}[ht!]
    \centering
    \includegraphics[width=\linewidth]{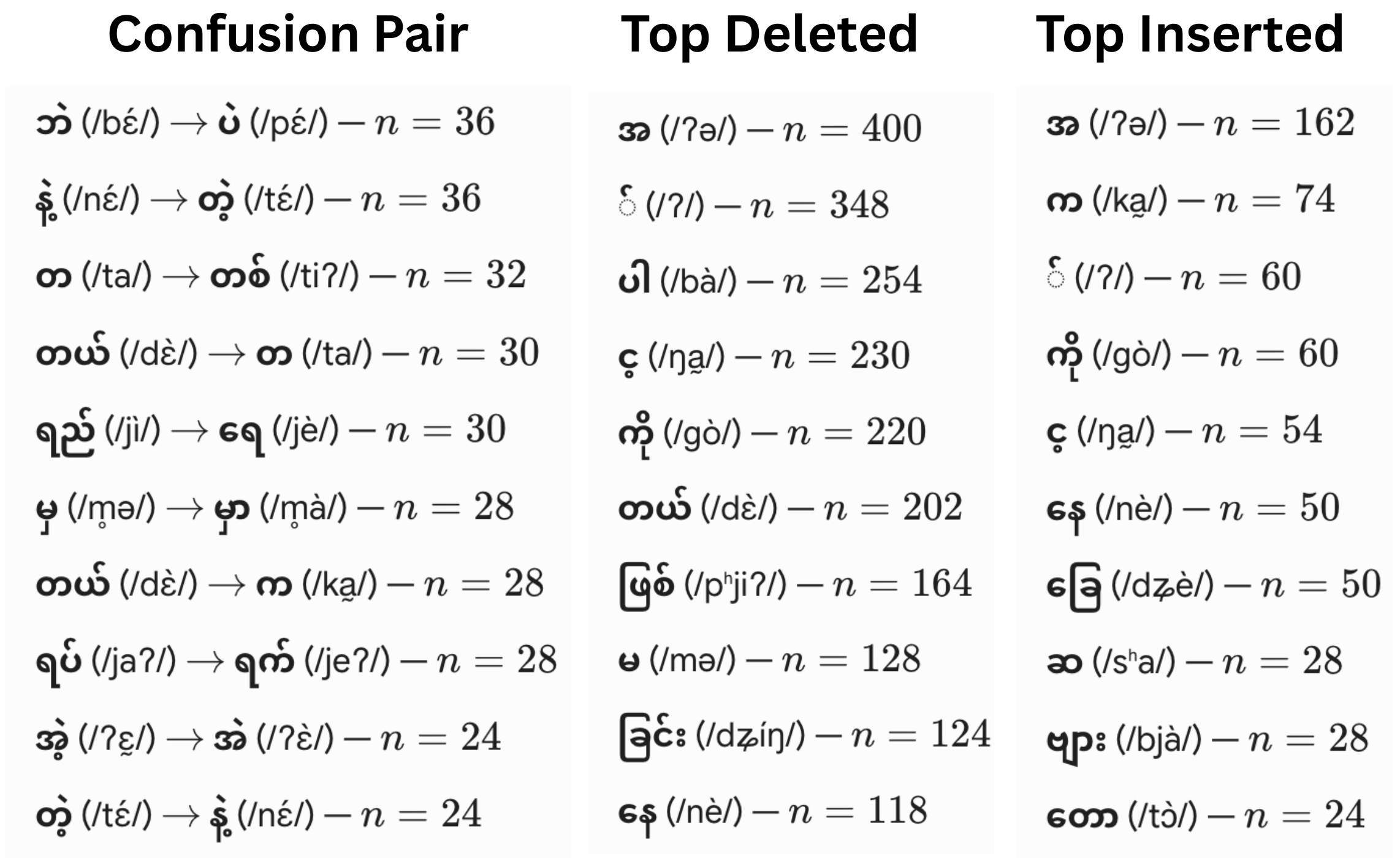}
    \caption{Syllable-level error breakdown for the myMediWhisper Medium Aug model.}
    \label{fig:error_analysis}
    \vspace{-3mm}
\end{figure}

The identified errors fall into three main categories: phonological and orthographic substitutions, function-word deletions, and weak-prefix insertions. Figure \ref{fig:error_analysis} presents the top 10 substitution confusion pairs ($\text{Ref} \rightarrow \text{Hyp}$), top 10 deleted syllables, and top 10 inserted syllables alongside their occurrence counts ($n$) and International Phonetic Alphabet (IPA) transcriptions. High error rates are concentrated in voicing/tonal confusions, as well as deletions of weak nominal prefixes (\textipa{/\textglotstop/}) and grammatical particles. 

\noindent \textbf{Phonological and Orthographic Substitutions.} Substitutions formed a primary source of alignment errors, driven by acoustic similarity, tone shifts, and colloquial orthographic variants. The most frequent substitution pair was \textipa{/b\'\textepsilon/} $\rightarrow$ \textipa{/p\'\textepsilon/} ($n = 36$), reflecting acoustic ambiguity between voiced and unvoiced bilabial plosives. Similar plosive-nasal confusion occurred between \textipa{/n\'\textepsilon/} $\rightarrow$ \textipa{/t\'\textepsilon/} ($n = 36$) and its reverse \textipa{/t\'\textepsilon/} $\rightarrow$ \textipa{/n\'\textepsilon/} ($n = 24$). Vowel height shifts and tone variations were also prevalent in postpositions and verbal markers, such as \textipa{/\textsubcircum{m}\textschwa/} $\rightarrow$ \textipa{/\textsubcircum{m}\~*a/} ($n = 28$), \textipa{/j\`i/} $\rightarrow$ \textipa{/j\`e/} ($n = 30$), and \textipa{/d\`\textepsilon/} $\rightarrow$ \textipa{/k\~a/} ($n = 28$). Furthermore, the model exhibited sensitivity to colloquial and transcriptive variations, frequently mapping creaky tone markers to low-tone informal variants as in \textipa{/\textglotstop\~{\textepsilon}/} $\rightarrow$ \textipa{/\textglotstop\`\textepsilon/} ($n = 24$), or replacing reduced forms with explicit numerals like \textipa{/ta/} $\rightarrow$ \textipa{/ti\textglotstop/} ($n = 32$) and full markers like \textipa{/d\`\textepsilon/} $\rightarrow$ \textipa{/ta/} ($n = 30$). Glottal vowel shifts were also observed (\textipa{/ja\textglotstop/} $\rightarrow$ \textipa{/je\textglotstop/} ($n = 28$)).

\noindent \textbf{High Deletion Rates in Function Words and Weak Syllables.} Deletions heavily outweighed insertions, primarily impacting unstressed initial syllables and grammatical markers. The unstressed nominalizing prefix \textipa{/\textglotstop @/} was the single most omitted token ($n = 400$), as initial weak syllables in continuous spoken Burmese lack strong acoustic energy peaks, making them highly vulnerable to acoustic masking. High-frequency functional items such as polite markers (\textipa{/b\`a/}, $n = 254$), object markers (\textipa{/\textg\`o/}, $n = 220$), realis verb markers (\textipa{/d\`{\textepsilon}/}, $n = 202$), progressive aspect markers (\textipa{/n\`e/}, $n = 118$), stative verbs (\textipa{/p\super{h}ji\textglotstop/}, $n = 164$), negative prefixes (\textipa{/m@/}, $n = 128$), and nominalizing suffixes (\textipa{/d\textctj\'{\i}\ng/}, $n = 124$) suffered frequent deletions during rapid speech. Additionally, isolated sub-syllabic diacritics like the glottal Asat marker (\textipa{/\textglotstop/}, $n = 348$) and nasal tone components (\textipa{/\ng\~*a/}, $n = 230$) frequently appeared in deletion statistics due to minor segmentation alignment boundaries.

\noindent \textbf{Spurious Insertions.} Spurious insertions were significantly lower than deletions and substitutions, demonstrating the model's strong resistance to hallucinating extended sequences. Inserted tokens were predominantly restricted to weak prefix additions like \textipa{/\textglotstop\textschwa/} ($n = 162$) or spurious postpositional and structural markers such as \textipa{/k\~*a/} ($n = 74$), \textipa{/\textg\`o/} ($n = 60$), \textipa{/\ng\~*a/} ($n = 54$), and \textipa{/n\`e/} ($n = 50$). Minor isolated fragments like \textipa{/d\textctj\`e/} ($n = 50$), \textipa{/s\super{h}a/} ($n = 28$), \textipa{/bj\`a/} ($n = 28$), and \textipa{/t\`{\textopeno}/} ($n = 24$), along with isolated diacritics like \textipa{/\textglotstop/} ($n = 60$), accounted for the remaining low-level insertion noise.

\section{Conclusion}
\label{sec:conclusion}

This work presents a comprehensive study of Burmese medical ASR for clinical dialogues utilizing large-scale Whisper architectures. By benchmarking domain-specific speech adaptation against external general-domain baselines trained on public assets like OpenSLR, we demonstrate that specialized fine-tuning is indispensable for capturing complex Burmese medical dictation. Our empirical evaluations establish that when hardware resources permit, full fine-tuning (FFT) delivers the highest precision, yielding a state-of-the-art Word Error Rate (WER) of 23.44\% with our \texttt{myMediWhisper-Medium} configuration. Conversely, Parameter-Efficient Fine-Tuning (PEFT) via high-rank Rank-Stabilized LoRA (rsLoRA, $r=128$) serves as a vital scaling mechanism, enabling the adaptation of Whisper-Large-v2 under strict memory constraints, albeit at the cost of increased RTF latency during inference. Furthermore, our exploration of multi-level data augmentation highlights a critical accuracy–robustness trade-off; while acoustic variations introduce minor regularization penalties on clean test sets, they substantially safeguard performance within highly noisy and reverberant environments. Complementing these quantitative metrics, our syllable-level error analysis reveals that residual model failures are predominantly driven by acoustic voicing overlaps and the deletion of weak nominal prefixes (\textipa{/\textglotstop/}) and grammatical particles due to low acoustic energy in continuous speech. Collectively, these insights establish an empirical foundation and provide explicit architectural guidance for deploying speech recognition frameworks inside realistic clinical environments.

\noindent \textbf{Limitations.}
Our robustness evaluation relies on simulated noise profiles and synthetic room impulse responses, which may not fully capture real-world clinical dynamics such as multi-speaker babble, non-stationary device noise, and microphone distortions. Additionally, despite verification by native speakers, our specialized dataset remains constrained in overall acoustic scale, speaker diversity, and regional dialect representation.

\noindent \textbf{Future Work.}
Future research will scale our corpus using an additional repository of collected clinical recordings, which are currently undergoing native-speaker validation to further balance gender distribution, expand dataset volume, and diversify medical specialties. With this extended validated corpus, we plan to broaden our research scope beyond automatic speech recognition toward downstream clinical speech tasks, including direct speech translation and medical named entity recognition (NER). Methodologically, we intend to deploy these real-world recordings to transition our evaluations from simulated robustness profiles to operational clinical metrics, while exploring alternative architectures and incorporating structural constraints targeting Burmese linguistic features—specifically tonal pitch contours and phonetic variations in localized medical loanwords.

\section*{Acknowledgements}

The authors would like to express their sincere gratitude to Hay Man Htun (Master Candidate, Kasetsart University / TAIST-Tokyo Tech) for generously sharing medical ASR data for this research. We deeply appreciate Hlaing Myat Nwe (Dual Doctoral Candidate, Japan Advanced Institute of Science and Technology (JAIST) and Sirindhorn International Institute of Technology (SIIT)) and Khaing Hsu Wai (Assistant Professor, Akita University, Japan) for their contributions to data recording. We are also profoundly grateful to our dedicated recording interns for their immense efforts: Sai Wai Yan Phyo, Thiri Thaw, Moe Chan Myae Maung, Thet Htet San, Myat Oo Swe, Su Sandi Linn, Htet Arkar, Min Thiha Tun, Saw Zi Dunn, Kyi Thant Sin, Htut Ko Ko, Eaint Lay Hmone, Htwe Myat Cho, Yadana Myint Hein, Ye Bhone Lin, Cham Myae Phyo, Thant Htut Aung, Nann Oak (Caesar), Kyawt Eaindray Win, Thet Su Sann, and Phoo Pwint Cho Thar. We explicitly note that while a subset of their recorded voice assets is not featured in the current iteration of this paper due to our rigorous, ongoing native-speaker validation process, their contributions are instrumental to the future expansion and scaling of this medical speech repository. The speech data and models will be released under \textbf{CC BY-NC-SA (Creative Commons Attribution-NonCommercial-ShareAlike)}.



\bibliography{anthology,custom}
\bibliographystyle{acl_natbib}




\end{document}